\documentclass{article}
\usepackage{amsmath}

\PassOptionsToPackage{numbers, compress}{natbib}

\usepackage[preprint]{neurips_2025}

\usepackage[utf8]{inputenc} 
\usepackage[T1]{fontenc}    
\usepackage{hyperref}       
\usepackage{url}            
\usepackage{booktabs}       
\usepackage{amsfonts}       
\usepackage{nicefrac}       
\usepackage{microtype}      
\usepackage{xcolor}         
\usepackage{threeparttable}

\usepackage{graphicx} 
\usepackage{subfigure}
\usepackage{multirow}

\title{Evidential Deep Active Learning for Semi-Supervised Classification}

\author{
\textbf{Shenkai Zhao}$^{1}$, \textbf{Xinao Zhang}$^{1}$, \textbf{Lipeng Pan}$^{1}$, \textbf{Xiaobin Xu}$^{2}$, \textbf{Danilo Pelusi}$^{3}$\\
$^1$College of Information Engineering, Northwest A\&F University, Yangling 712100, Shanxi, China \\
$^2$Hangzhou Dianzi University, Hangzhou 310018, China \\
$^3$Department of Communications Sciences, University of Teramo, Italy \\
\texttt{jisuanjizsk@nwafu.edu.cn}, \texttt{zhangxinao1998@163.com}, \texttt{lipeng.pan@nwafu.edu.cn},\\ 
\texttt{xuxiaobin1980@hdu.edu.cn}, \texttt{dpelusi@unite.it}
}

\begin{document}


\maketitle

\begin{abstract}

Semi-supervised classification based on active learning has made significant progress,  but the existing methods often ignore the uncertainty estimation (or reliability) of the prediction results during the learning process, which makes it questionable whether the selected samples can effectively update the model. Hence, this paper proposes an evidential deep active learning approach for semi-supervised classification (EDALSSC). EDALSSC builds a semi-supervised learning framework to simultaneously quantify the uncertainty estimation of labeled and unlabeled data during the learning process. The uncertainty estimation of the former is associated with evidential deep learning, while that of the latter is modeled by combining ignorance information and conflict information of the evidence from the perspective of the T-conorm operator. Furthermore, this article constructs a heuristic method to dynamically balance the influence of evidence and the number of classes on uncertainty estimation to ensure that it does not produce counter-intuitive results in EDALSSC.  For the sample selection strategy, EDALSSC selects the sample with the greatest uncertainty estimation  that is calculated in the form of a sum when the training loss increases in the latter half of the learning process. Experimental results demonstrate that EDALSSC outperforms existing semi-supervised and supervised active learning approaches on image classification datasets. \footnote{Code is available at \url{https://github.com/EDALSSC/EDALSSC}}

           
\end{abstract}

\section{Introduction}

Remarkable success of deep learning \cite{lecun2015deep, goodfellow2016deep} in a wide variety of classification tasks heavily relies on large amounts of labeled data. However, the acquisition of labels for samples is time-consuming, labor-intensive, and often requires domain-specific expertise in various tasks, such as in the medical field \cite{budd2021survey} and autonomous driving \cite{hekimoglu2024monocular}. In response to this challenge,  some artistic methods and strategies have been proposed one after another, such as
semi-supervised learning\cite{zhu2005semi,van2020survey}
and 
active learning \cite{settles2009active,ren2021survey}.
The former utilizes the structural information of unlabeled samples to guide or constrain the model, so as to obtain more stable and generalized decision boundary under the condition of a small number of labels. While the latter enables the model to actively select the most valuable samples for annotation, thereby reducing the annotation cost and optimizing the decision boundary.
Semi-supervised learning and active learning alleviate the scarcity problem of labeled samples from different perspectives. This inspired semi-supervised learning and active learning to complement each other in forms to avoid the waste of labels in classification tasks \cite{sinha2019variational,zhang2020state, gao2020consistency, guo2021semi,huang2021semi,zhang2022boostmis,zhu2003combining,zhou2013active}.

\begin{figure}[htbp]
  \centering

  \subfigure[Shannon entropy and Ignorance in CIFAR-10]{
    \includegraphics[width=0.3\textwidth]{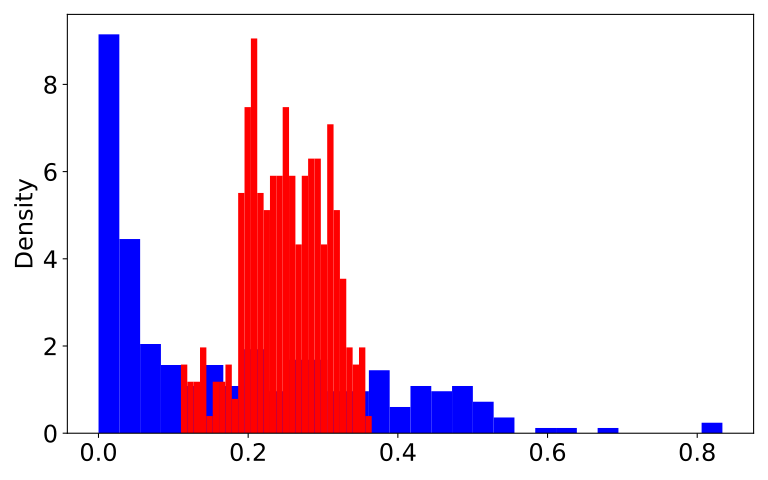}
    \label{fig:uncertainty_cifar10}
  }
  \quad
  \subfigure[Shannon entropy and Conflict in CIFAR-10]{
    \includegraphics[width=0.3\textwidth]{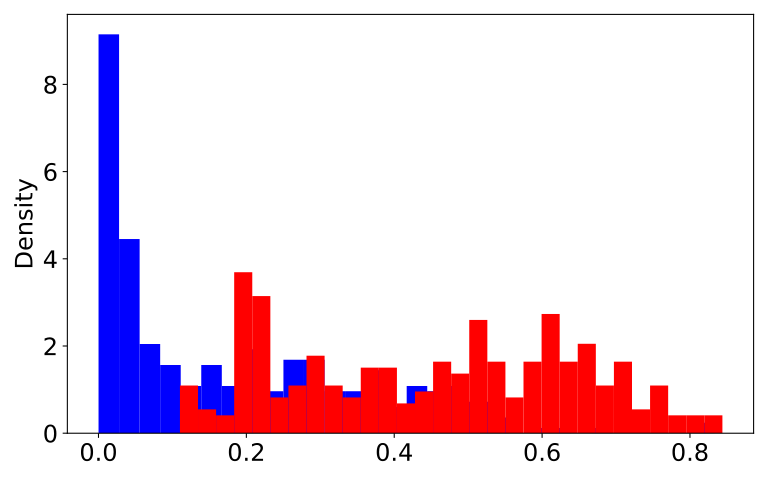}
    \label{fig:conflict_cifar10}
  }
  \quad
  \subfigure[Shannon entropy and  Uncertainty estimation in CIFAR-10]{
    \includegraphics[width=0.3\textwidth]{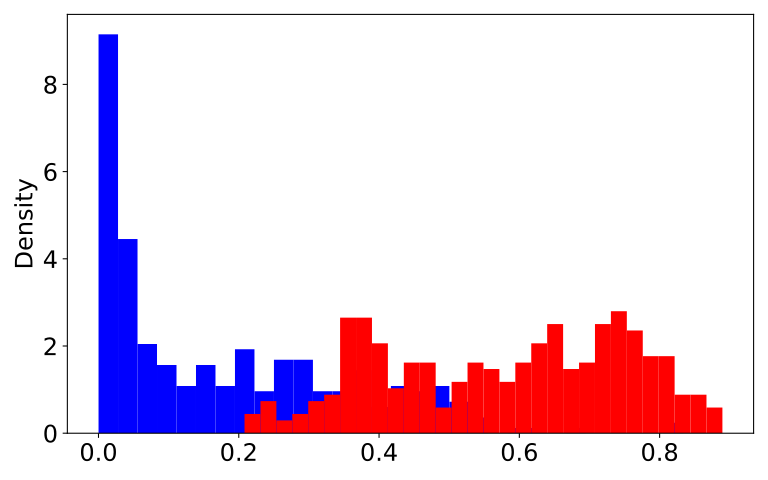}
    \label{fig:fusion_cifar10}
  }

  \vspace{0.5cm}  
  
  \subfigure[Shannon entropy and  Ignorance in CIFAR-100]{
    \includegraphics[width=0.3\textwidth]{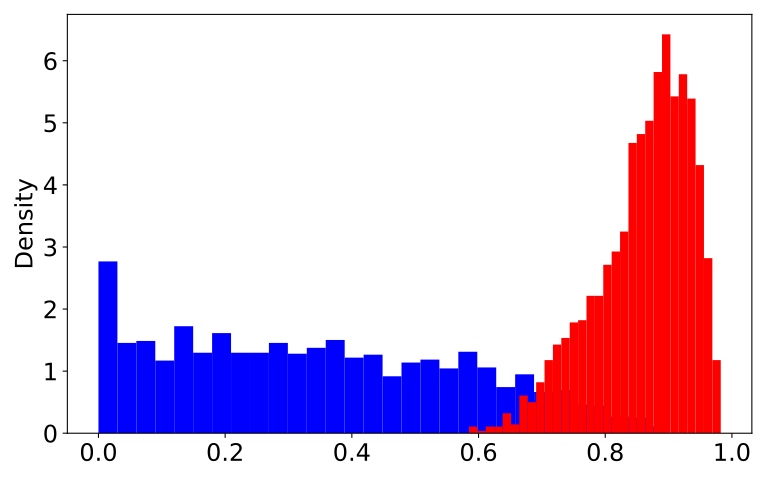}
    \label{fig:uncertainty_cifar100}
  }
  \quad
  \subfigure[Shannon entropy and  Conflict in CIFAR-100]{
    \includegraphics[width=0.3\textwidth]{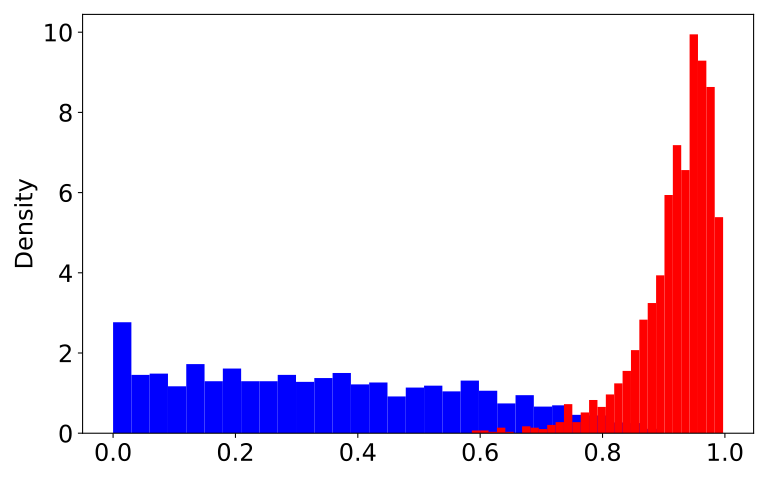}
    \label{fig:conflict_cifar100}
  }
  \quad
  \subfigure[Shannon entropy and  Uncertainty estimation in CIFAR-100]{
    \includegraphics[width=0.3\textwidth]{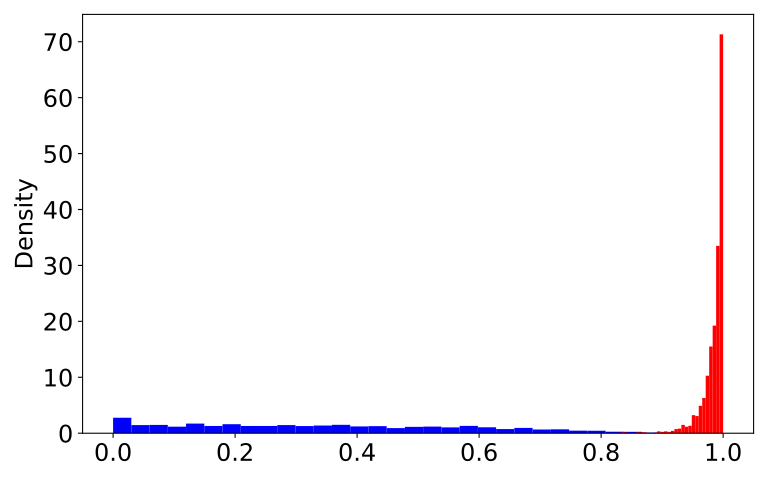}
    \label{fig:fusion_cifar100}
  }

  \caption{Entropy of the prediction result misclassify the same samples in CIFAR-10 and CIFAR-100. \textbf{Blue:} Shannon entropy for SSCAL. \textbf{Red:} Ignorance, Conflict and Uncertainty estimation for the proposed method.}
  \label{fig:distribution_comparison}
\end{figure}

However, the existing semi-supervised classification with active learning (SSCAL) methods  \cite{huang2021semi} ignore the uncertainty estimation of the sample prediction results, making the model overly confident, as shown in Figure \ref{fig:distribution_comparison}. Figure \ref{fig:distribution_comparison} shows the uncertainty of the prediction results for the two models on the misclassified samples. For SSCAL, its uncertainty (calculated through normalization of  the Shannon entropy) is low on the misclassified samples, which indicates that SSCAL has a high level of confidence in the prediction results of the samples, even if in fact the sample is misclassified by the model (the model does not know whether the sample is misclassified). For our method, its uncertainty is high on the misclassified  samples (in fact, the model does not know whether this sample is misclassified or not.) This indicates that when the uncertainty of the sample prediction results is high, the samples may be misclassified. Compared with the proposed method, the sample prediction results of SSCAL are overly confident. The reason lies in the lack of uncertain estimation in the training stage and the active learning stage of SSCAL. Therefore, it is questionable whether the high-value samples selected by the overly confident SSCAL can effectively update the model, as shown in Table \ref{tab:cv_parameter_diff}. Table \ref{tab:cv_parameter_diff} lists the coefficient of variation of the parameter space difference of the model before and after sample selection. The greater the parameter space difference coefficient is, the higher the update degree of the model will be. It can be seen from Table \ref{tab:cv_parameter_diff} that, compared with the proposed method, the sample selection of SSCAL does not effectively update the model. That is to say, SSCAL is unable to select samples of high value. Therefore, it is imperative to introduce uncertain estimation into the training stage and sample selection stage of semi-supervised active learning.


\begin{table}[htbp]
\centering
\begin{threeparttable}
\caption{Coefficient of Variation (CV) of parameter space difference Between Consecutive cycles}
\label{tab:cv_parameter_diff}  
\begin{tabular}{llcccccc}
\toprule
\textbf{Dataset} & \textbf{Method} & $\Delta$1--2 & $\Delta$2--3 & $\Delta$3--4 & $\Delta$4--5 & $\Delta$5--6 & $\Delta$6--7 \\
\midrule
\multirow{2}{*}{CIFAR10} 
& TOD & 615.4972 & 643.9383 & 662.8177 & 677.8622 & 689.6953 & 699.9733 \\
& \textbf{Ours} & \textbf{627.9706} & \textbf{660.2504} & \textbf{680.6517} & \textbf{699.5919} & \textbf{713.5056} & \textbf{724.9323} \\
\midrule
\multirow{2}{*}{CIFAR100} 
& TOD & 569.4792 & 636.3138 & 654.2925 & 668.1279 & 677.8381 & 685.7693 \\
& \textbf{Ours} & \textbf{588.2388} & \textbf{650.4489} & \textbf{685.7529} & \textbf{705.0800} & \textbf{718.8955} & \textbf{731.4552} \\
\bottomrule
\end{tabular}
\begin{tablenotes}
\footnotesize
\item Note:Parameter space difference $\left ( \left\|\theta _{t+1}-\theta _{t} \right\| \right )$ is the indicator for model update \cite{bryan2005active}. To comprehensively display the update degree of the model after adding samples, this paper uses the coefficient of variation of Parameter space difference instead of Parameter space difference. $\Delta$t-t+1  represents the coefficient of variation of $\left ( \left\|\theta _{t+1}-\theta _{t} \right\| \right )$ between the model for the \textit{t-th} cycle and the model  for the \textit{(t+1)-th}  cycle in active learning. $\theta _{t}$ represents the parameters of the model in the \textit{t-th} cycle of active learning. The higher the value of the Coefficient of Variation is, the higher the update degree of the model will be.
\end{tablenotes}
\end{threeparttable}
\end{table}

Based on the above discussion, this paper proposes an  evidential deep active learning approach for semi-supervised
classification (EDALSSC). In EDALSSC, with the aid of the uncertainty-aware mechanism, this article considers the uncertainty estimation of labeled samples and unlabeled samples in the learning stage.  The uncertainty estimation of labeled samples is calculated through the cross-entropy loss of evidence, while the uncertainty estimation of unlabeled samples is obtained by  combining  the ignorance information and conflict information of combined evidence in the form of \textit{T-conorm}. For EDALSSC, this article designs a heuristic method to dynamically balance the influence of evidence and the number of classes on ignorance information and conflict information. Moreover, for each cycle of active learning, EDALSSC selects the samples with the greatest sum of uncertainty estimation when the loss increases in the second half of the training stage.  Finally, EDALSSC is applied to the image dataset experiments. The experimental results show that EDALSSC is superior to other semi-supervised active learning and supervised active learning methods.
In summary, the contributions of this paper are as follows:

(1) This article introduces the uncertainty estimation of labeled and unlabeled samples into semi-supervised active learning, so as to avoid the model being overly confident and generating counterintuitive results.

(2) EDALSSC quantifies the uncertainty estimation of unlabeled samples by combining ignorance information and conflict information with the T-conrom operator, aiming to provide reliable indicators for model training and sample selection.

(3) EDALSSC introduces a dynamic Dirichlet density parameter scaling mechanism from a heuristic perspective, in order to balance the influence of evidence and the number of categories on ignorance information and conflict information.

\section{Preliminaries}
\textbf{Evidential deep learning (EDL) \cite{sensoy2018evidential}:} 
In the framework of subjective logic or evidence theory, Sensoy \textit{et al.} use the Dirichlet density parameter to model the uncertainty estimation of the prediction results, enabling the model to say \textit{"I don't know."} 
In evidence theory, each of the $K$ mutually exclusive singletons is assigned a subjective opinion $b_k$, along with an overall uncertainty mass of $u$. These components satisfy the normalization constraint:

\begin{equation*}
\sum_{k=1}^{K} b_k + I = 1
\end{equation*}

where $I \geq 0$ and $b_k \geq 0$ for$ \quad k = 1, \ldots, K$.  The subjective opinions $b_k$ and uncertainty $u$ are then computed as follows:  

\begin{equation*}
b_k = \frac{e_k}{S}, \quad I = \frac{K}{S},
\end{equation*}

where $S = \sum_{i=1}^{K} (e_k + 1)$.
The Dirichlet distribution is a multivariate probability distribution defined over probability vectors. It is commonly used to model the probabilities of a set of mutually exclusive and collectively exhaustive events. The Dirichlet distribution is parameterized by a vector $\boldsymbol{\alpha} = (\alpha_1, \dots, \alpha_K)$, and its probability density function  is defined as:
\[
\text{Dir}(p|\boldsymbol{\alpha}) = 
\begin{cases}
\frac{1}{B(\boldsymbol{\alpha})} \prod_{i=1}^{K} p_i^{\alpha_i - 1}, & \text{if } p \in S_K, \\
0, & \text{otherwise},
\end{cases}
\]
where 
$B(\boldsymbol{\alpha})$ is the multivariate Beta function, and  \( S_K \) is the simplex.


\textbf{Semi-supervised Active Learning:}
Some studies have adopted a complementary form of semi-supervised learning and active learning to alleviate the problem of sample label scarcity.
 Huang \textit{et al.} employe Temporal Output Discrepancy (TOD) \cite{huang2021semi} to estimate model loss and select samples with the greatest difference of model predictions between the current and previous cycle, thereby improving the model’s ability to learn from highly uncertain samples.
 Sinha \textit{et al.} propose a VAE-GAN structure (VAAL) that jointly learns latent representations of labeled and unlabeled data \cite{sinha2019variational}. The discriminator undergoes adversarial training to determine whether a sample belongs to a labeled pool or an unlabeled pool.  
 Gao \textit{et al.}  formalize a method that trains models using enhanced consistency strategies, and selects the most inconsistent subset of predictions from a set of random augmentations applied to each given sample\cite{gao2020consistency}. 
 Guo \textit{et al.} introduce a method based on graph propagation of labeled semantic information to generate pseudo-labels for unlabeled samples, and used virtual adversarial perturbations to identify boundary samples, followed by entropy-based sample selection\cite{guo2021semi}. 
Although these most advanced methods have made significant progress, they ignore the reliability of model predictions and make it difficult to select high-value samples for training. Therefore, this article proposes the evidential deep active learning approach for semi-supervised classification.

\section{Semi-Supervised Evidential Deep Active Learning}

\begin{figure}[htbp]
  \centering
  \includegraphics[width=1\textwidth]{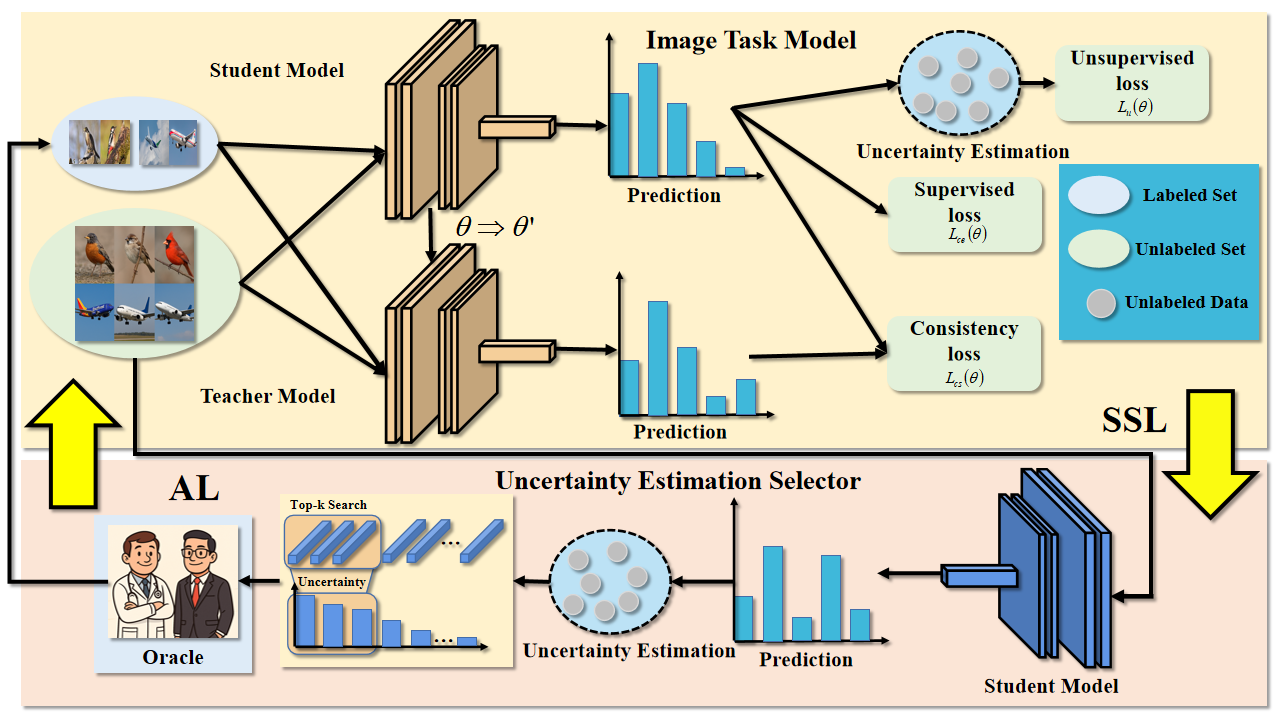}
  \caption{\textbf{Overview of EDALSSC.} It consists of two modules: (a) \textit{Image Task Model} trains the base model jointly with labeled and unlabeled data, optimizing a composite objective that combines evidence-based cross entropy loss on labeled samples, uncertainty estimation (using T-conorm to comprehensively consider ignorance information and conflict information) on unlabeled samples, and consistency loss across all samples; (b) \textit{Uncertainty Estimation Selector} selects supplementary samples with higher uncertainty estimates, which have higher total uncertainty estimation during the middle-to-late stages of training, when training loss increases. 
}
  \label{fig:lct02}
\end{figure}

This section considers the problem of semi-supervised active learning under the presence of both labeled and unlabeled data. Let $\mathcal{D} = \mathcal{L} \cup \mathcal{U}$ denote the entire dataset, where $\mathcal{L} = \{(x_i, y_i)\}_{i=1}^{n_l}$ is a small labeled set and $\mathcal{U} = \{x_j\}_{j=1}^{n_u}$ is a large pool of unlabeled samples, with $n_u \gg n_l$. Our goal is to train a classification model $f_\theta: \mathcal{X} \rightarrow \mathcal{Y}$ that generalizes well by iteratively selecting small batches of informative samples from  $\mathcal{U}$ to query for labels. Furthermore, the overall framework of EDALSSC is shown in Figure \ref{fig:lct02},
and detailed work is as shown in the following subsections.

\subsection{Dynamically  scaling of Dirichlet density parameters}

Ignoring the balance between evidence and the number of class can lead to counterintuitive results of uncertainty estimation in EDALSSC.
For  a 100-classification problem, its evidence is represented as follows,
\( \mathbf{e} = [100, 0, 0, 0, \ldots, 0] \),
that is to say, the evidence for all channels except the first one is $0$. In this scenario, the prediction result of the sample should be appropriate, but its uncertainty estimation is as high as one half, which is the counterintuitive result of uncertainty estimation caused by the overly conservative model. 
Thus, we introduce a \textit{dynamic scaling factor} $r$ from a heuristic perspective. This factor is computed by considering the discrepancy between the model's prediction scores for the most probable and the second most probable classes.

\begin{equation*}
r = 
\begin{cases}
1, & \text{if } e_{\text{max}}^2 + e_{\text{second}}^2 = 0 \\
\frac{(e_{\text{max}} + e_{\text{second}})^2}{2 \cdot (e_{\text{max}}^2 + e_{\text{second}}^2)}, & \text{otherwise}
\end{cases}
\end{equation*}
Adjusting the Dirichlet density parameter is a method of relaxing constraints, and its rationality has been proven \cite{chen2024r}. The adjusted Dirichlet density parameters are as follows.


\[
\alpha_{ik} = e_{ik} + 1 \cdot r
\]
then
\[
S_i = \sum_{k=1}^{K} \alpha_{ik} = \sum_{k=1}^{K} (e_{ik} + 1 \cdot r),
\quad
I_i = \frac{K\cdot r}{S_i}
\]
where $K$ denote the number of classes.
As an heuristic method, when $I_i$ reaches its maximum value (or $ r = 1$), it corresponds to two different evidence distributions. (1) The first distribution is that all the evidence is 0. That is to say, none of the evidence supports the corresponding proposition. This provides a kind of global ignorance information, corresponding to  $I_i$ = 1. 
(2) The other case is when $e_{max} = e_{second}$, that is, when the evidences for the two propositions with the highest support are equal, the model still cannot provide any effective information for decision-making. That is to say,  $r$ = 1. 
By adaptively scaling $\boldsymbol{\alpha}$ based on the balance of the number of class and the evidence, which enhances the discriminative power of uncertainty estimation, leading to more reliable uncertainty estimation for samples (The ablation experiment results related to this work were presented  by \textit{Ablation 2} in Table \ref{tab:ablation-study}).

\subsection{Uncertainty Decomposition and Aggregation}
\label{sec:uncertainty}
For EDALSSC, uncertainty estimation plays an important role in the loss function and sample selection strategy. Therefore, the importance of setting a reasonable uncertainty estimation method is self-evident. In evidence theory, uncertainty includes \textit{ignorance} and \textit{conflict}. Ignorance is related to the evidence supporting multi-element propositions (especially global focal elements) , while conflict is related to the evidence supporting different single-element propositions.

\textbf{\textit{Ignorance:}} In EDALSSC, $I_i$ can naturally express the ignorant information in the evidence. Obviously, $I_i$ satisfies the following properties. (1) $ I_{i}\in \left [ \frac{max\left\{ e_{k}\right\}}{max\left\{ e_{k}\right\}+K},1 \right ]$.
When all the evidence is 0, $I_i$=1. If and only if exactly one piece of evidence is non-zero,  $u_i = \frac{max\left\{ e_{k}\right\}}{max\left\{ e_{k}\right\}+K}$. (2) If $e_{j}\sqsubseteq _{pl}e_{t}$ or $e_{j}\sqsubseteq _{q}e_{t}$, then $I_t \geq I_l$, where $\sqsubseteq _{pl}$ is $pl-$ordering, and $\sqsubseteq _{q}$ is $q-$ordering \cite{zhou2023information}.

\textbf{\textit{Conflict:}} In EDALSSC, conflict is generally used to quantify the degree of dispersion among the propositional support degrees within evidence \cite{barhoumi2025empirical}. Inspired by the linear relationship between $u_i$ and the evidence, the conflict is calculated as follows
from the perspective of linear relationships.
\begin{equation*}
C_i = 1 - \frac{B_i}{K-1}
\end{equation*}
where $B_{i}=\sum_{k=1}^{K}\left ( b_{i,max} -b_{i,k}\right )$, $b_{i,max}=Max(b_{i,k})$, $b_{i,k}$ denote the belief mass assigned to the $k$-th class for the $i$-th sample which is computed as $b_{i}=e_{i}/S_i$, $S_i = \sum_{k=1}^{K} \alpha_{ik} = \sum_{k=1}^{K} (e_{ik} + 1 \cdot r)$, $e_{ik}$ represents the evidence associated with class $k$ for the $i$-th sample, obtained from the output of the convolutional neural network.
By extracting the maximum belief quality of the sample from the model and calculating the total difference between it and the other belief qualities. If the sum of the differences is large, it indicates that the model's prediction results for the current sample are relatively clear, that is, not conflicting. On the contrary, if the total difference is small, it indicates that the belief distribution of the model between different calss is relatively close, and there is a high degree of conflict in the prediction. Furthermore, $C_{i}\in \left [ 0,1 \right ]$. When $b_{ik}=1$ and $b_{ij}=0$ for $j \neq k$, then $C_{i}=0$. When $b_{ik}=\frac{1}{x}$for$ \quad k = 1, \ldots, K$, then $C_{i}=0$. The calculation method of conflict information ensures that it is independent of ignorance information, which is also consistent with the meanings of both, providing a basis for information aggregation.

\textbf{\textit{Uncertainty estimation:}} In EDALSSC, ignorance and conflict have the same position on uncertainty estimation. That is to say, for both, as long as one of them reaches the maximum value, the corresponding sample needs to be paid attention to. Therefore, this article achieves the aggregation of ignorance and conflict from the perspective of \textit{OR} \cite{zhou2022modeling}, is described as follows.
\begin{equation*}
u_i = T\left(I_i, C_i\right)=1-\left(1-I_i\right) \times \left(1- C_i\right)= I_i  +  C_i  - I_i \times C_i
\end{equation*}
It should be pointed out that the aggregation of $I_i$ and $C_i$ satisfies the following properties. (1) $T(I_i, C_i) = T(C_i, I_i)$. 
(2) If  $I_i \leq I_{i}^{'},\ C_i \leq C_{i}^{'}$, then  $T(I_i,C_i) \leq T(I_{i}^{'}, C{i}^{'})$.
(3) $T(I_i , 0) = a,\quad T(I_i  , 1) = 1$  or  $T(C_i, 0) = a,\quad T(C_i , 1) = 1$.
(4) ) $T\left(I_i, C_i\right)\in \left [ 0,1 \right ]$, when$ I_i=C_i=0$, $T\left(I_i, C_i\right)=0$. When $I_i$=1 or $C_i$=1, $T\left(I_i, C_i\right)=1$.   The ablation experiment results related to this work were presented  by \textit{Ablation 3} in Table \ref{tab:ablation-study}.

\subsection{Sample selection strategy}
\label{sec:strategy}
As the core component of active learning, the design of the sample selection strategy critically affects the quality and utility of the samples chosen for subsequent model training. Traditional semi-supervised active learning approaches typically overlook the notion of uncertainty estimation when selecting unlabeled samples. To address this, we leverage the  uncertainty estimation $u_i$ proposed in Section \ref{sec:uncertainty}, which integrates both ignorance and conflict, as the basis for selection.

Unlike the existing methods that estimate uncertainty only after the model has fully converged, our method calculates the sum of the  uncertainty estimation of the samples when the loss of the model increases in the second half of each cycle. Then sort them and select the samples with great uncertainty estimation in batches. The reason lies in the fact that the early training is not reliable. Therefore, attention should be paid to the second half of the model training stage. Furthermore, focusing on epochs with increased losses can prevent the model from being overly conservative and thereby achieve high classification performance (The ablation experiment results related to this work were presented  by \textit{Ablation 4} in Table \ref{tab:ablation-study}).


\subsection{Learning criterion}
To fully exploit the unlabeled data and maintain consistency with our sample selection strategy, EDALSSC introduces a uncertainty-aware unsupervised loss. Specifically, this subsection estimates the reliability of each unlabeled sample using the uncertainty score derived from both ignorance
 and conflict (as discussed in Section \ref{sec:uncertainty}). 
Therefore, unsupervised loss are described as follows:
\begin{equation*}
U = \frac{1}{|\mathcal{U}|} \sum_{x_i \in \mathcal{U}} u_i
\end{equation*}

Moreover, this paper incorporate consistency regularization \cite{laine2016temporal,tarvainen2017mean}, a widely adopted principle in semi-supervised learning, is described as follows.
\begin{equation*}
L_{CS}(\theta) =\frac{1}{|\mathcal{L\cup U}|} \sum_{x_i \in \mathcal{(L\cup U)}} \left( \text{CS}\left[ f(x_i; \theta_1), f(x_i; \theta_2) \right] + \text{CS}\left[ f(x_i; \theta_2), f(x_i; \theta') \right] \right)
\end{equation*}
where $\theta _{t}^{'}=\omega \theta _{t-1}^{'}+\left ( 1-\omega  \right )\theta _{t}$, and  $\text{CS}(a, b) = \beta_1 \cdot  \text{MSE}(a, b) + \beta_2\cdot \text{KL}(a, b) $ denotes the consistency loss. In most cases, $\beta_1$ and $\beta_2$ are empirically set to 0.5. Here, $\text{MSE}$ and $\text{KL}$ refer to the mean squared error and Kullback–Leibler divergence, respectively, which are employed as complementary metrics to quantify the prediction consistency between different model outputs.

For labeled data, EDALSSC adopts evidence-based cross entropy loss, which encourages the model to produce accurate predictions on cleanly annotated samples. Given a batch of labeled examples $(x, y) \in \mathcal{L}$, the supervised loss is given by:
\begin{equation*}
\mathcal{L}(\Theta) 
= \sum_{x_i \in \mathcal{L}} \sum_{j=1}^K y_{ij} \left[ \psi(S_i) - \psi(\alpha_{ij}) \right]
\end{equation*}

where  $\psi(\cdot)$  is the digamma function, $S_i = \sum_{i=1}^{K} \alpha_i = \sum_{i=1}^{K} (e_k + 1 \cdot r)$.

To sum up, the overall training objective of EDALSSC combines both supervised and unsupervised losses, with a dynamic weighting scheme to gradually introduce unlabeled data into the learning process. The total loss is defined as:
\begin{equation*}
L_{overall}^{c}=L_{CE}^{c}\cdot factor +\lambda \cdot (U^{c}+L_{CS}^{c})
\end{equation*}
where $\lambda$ is a balancing coefficient set to 0.05 to govern the trade-off between the task-specific objective and the unsupervised regularization term.
The term $ \textit{factor} = 1-\frac{\textit{cycle}}{\textit{num\_cycle}} $ is a decay factor associated with the progression of active learning cycles, where $\textit{cycle}$ denotes the current iteration index, and $\textit{num\_cycle}$ denotes the total number of active learning iterations. As training progresses, $\textit{factor}$ gradually decreases, enabling adaptive emphasis on different components of the learning objective. The ablation experiment results related to this work were presented  by \textit{Ablation 1} in Table \ref{tab:ablation-study}.

\section{Experiments}

\label{Experiments}

To systematically evaluate the effectiveness of EDALSSC, this conduct experiments on several benchmark image classification datasets. This section   presents the datasets, baseline methods, and implementation details, followed by the experimental results and a detailed analysis of performance in  datasets.

\subsection{Experimental setup}

\noindent\textbf{Dataset} This paper conducts experiments on four widely used benchmark datasets for image classification: CIFAR-10 \cite{krizhevsky2009learning}, CIFAR-100 \cite{krizhevsky2009learning}, SVHN \cite{netzer2011reading} and Fashion-MNIST \cite{xiao2017fashion}. CIFAR-10, SVHN and Fashion-MNIST consist of 10 calsses and serve as representatives of low-class-number settings, while CIFAR-100 contains 100 calsses, representing high-class-number scenarios. These datasets are widely adopted in the evaluation of models for image classification, semi-supervised learning, and active learning.

\noindent\textbf{Baseline} For a more comprehensive evaluation, this section conducts comparisons between our proposed framework and several state-of-the-art methods, including TOD\cite{huang2021semi}, CoreGCN\cite{caramalau2021sequential}, UncertainGCN\cite{caramalau2021sequential}, VAAL\cite{sinha2019variational}, Core-set\cite{sener2017active}, and lloss\cite{yoo2019learning}. Furthermore, random sampling (Random) and the model is trained on full datasets ("Full Training")  are adopted as the baselines for performance reference. As a straightforward baseline, this section  also directly employ the uncertainty  derived from the EDL framework to guide sample selection, prioritizing samples with higher uncertainty as the most informative candidates for labeling. Meanwhile, in the design of learning criteria, this section only considered the cross entropy and consistency loss under traditional semi-supervised tasks. Based on the above two perspectives, this section  designs a \textit{full ablation study} to investigate the advantages of our proposed EDALSSC over a naive combination of EDL and AL in Semi-Supervised Classification.

\noindent\textbf{Implementation details} For EDALSSC , this section adopts ResNet-18 \cite{he2016deep} as the backbone network. To initialize the model, this section  randomly selects sample 10\% of the dataset as labeled data to form the initial training set. An active learning strategy is then employed in an iterative manner. At each iteration, this section computes sum of uncertainty estimation of samples in the unlabeled pool, and select the top 5\% most valuable samples for annotation. These newly labeled samples are subsequently added to the training set to update the model.   Next, active learning  is repeated for 6 cycles. Furthermore, all the experiments are repeated three times. Further details can be found in the appendix.



\subsection{Experimental Results}
\begin{figure}[htbp]
  \centering
  \subfigure{
    \includegraphics[width=0.45\textwidth]{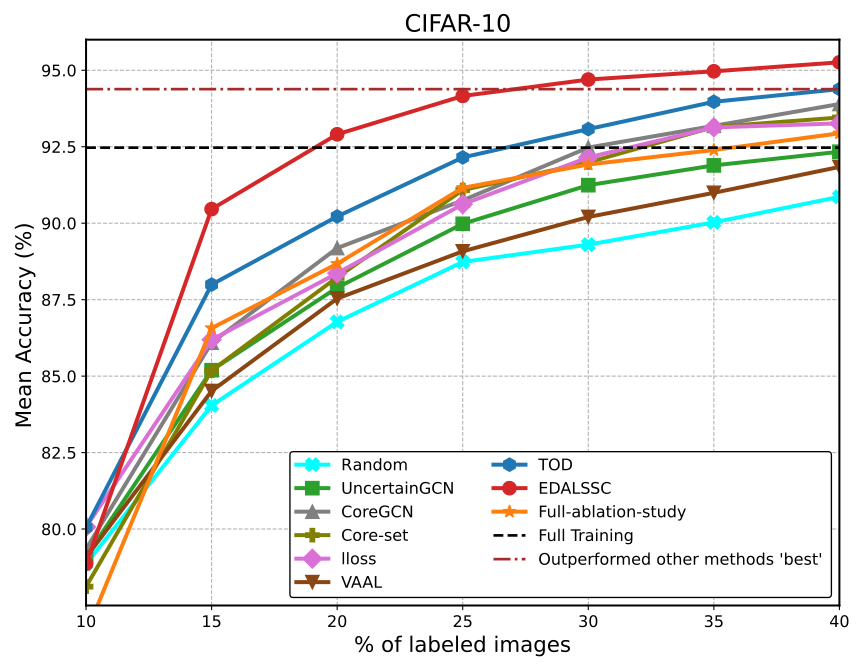}
    \label{fig:cifar10}
  }
  \hspace{0.05\textwidth}
  \subfigure{
    \includegraphics[width=0.45\textwidth]{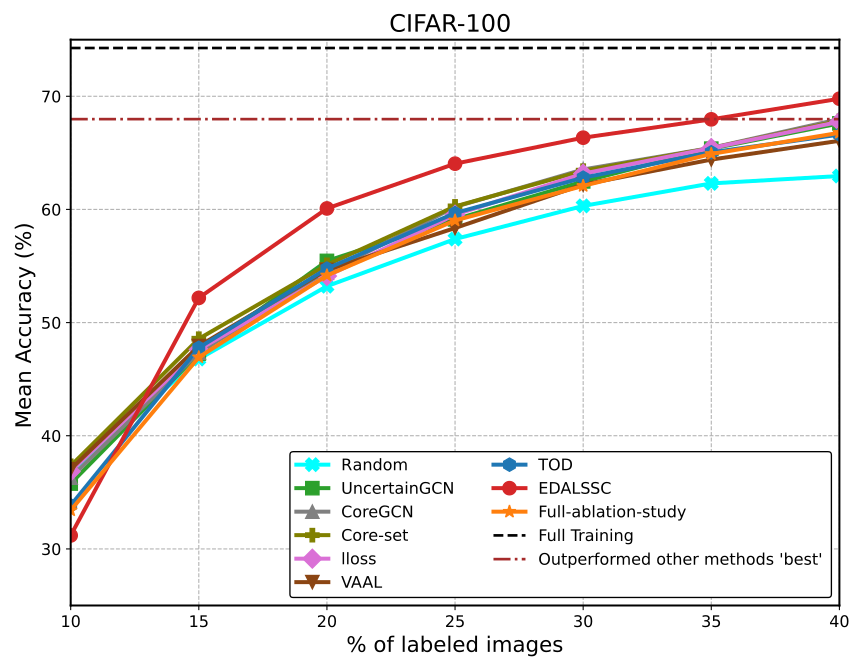}
    \label{fig:cifar100}
  }
  \subfigure{
    \includegraphics[width=0.45\textwidth]{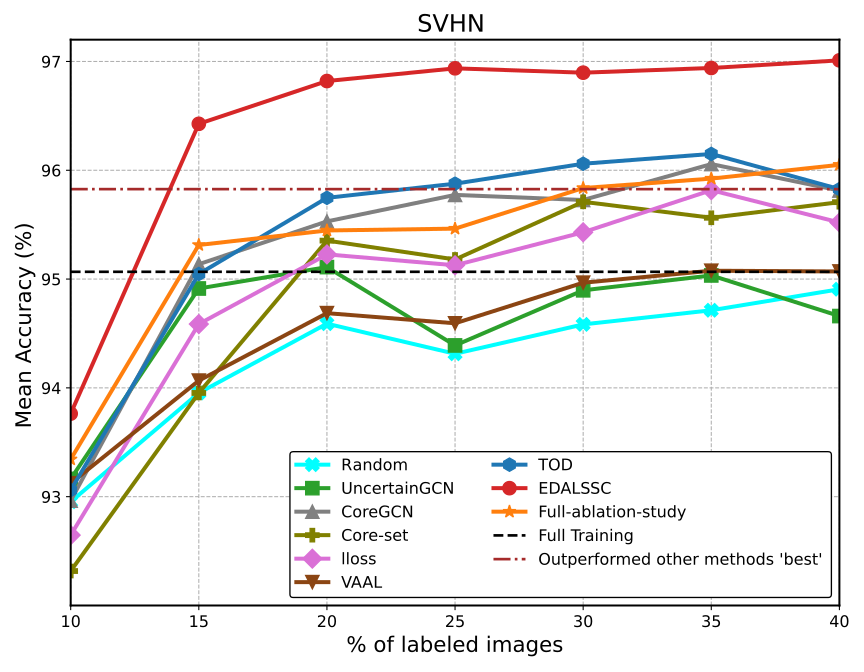} 
    \label{fig:svhn}
  }
  \hspace{0.05\textwidth}
  \subfigure{
    \includegraphics[width=0.45\textwidth]{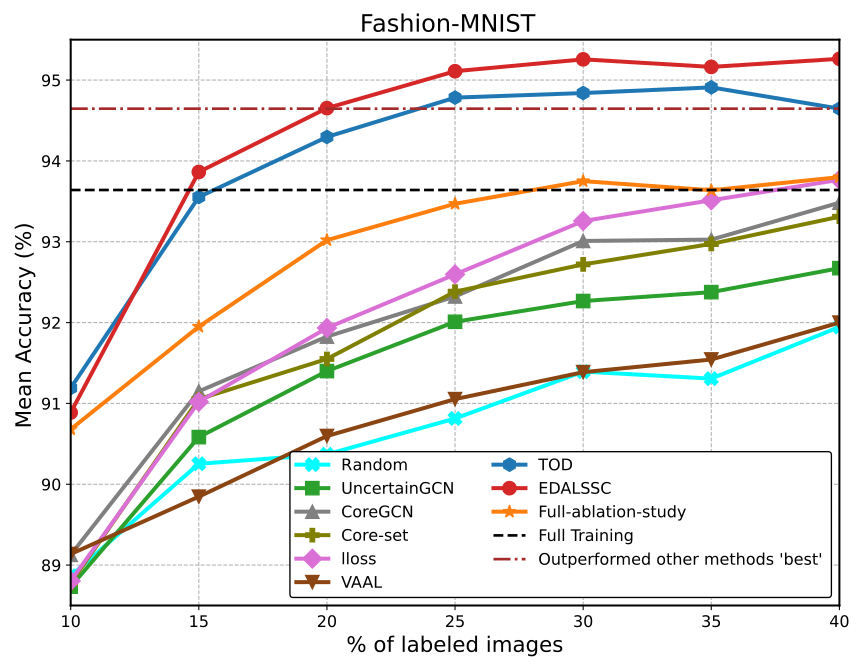} 
    \label{fig:Fashion-MNIST}
  }
  \caption{Performance comparison of image classification on four benchmark datasets.}
  \label{fig:Performance comparison of image classification on four benchmark datasets}
\end{figure}

For the four datasets, the experimental results of EDALSSC and the different methods mentioned in \textbf{Baseline} are shown in Figure \ref{fig:Performance comparison of image classification on four benchmark datasets}. It can be seen from Figure \ref{fig:Performance comparison of image classification on four benchmark datasets} that EDALSSC
consistently outperforms the compared approaches throughout the active learning process, except for the initial cycle. This consistently superior performance demonstrates the effectiveness of evidential deep active learning  for semi-supervised classification tasks.
In particular, for the CIFAR-10 dataset, EDALSSC reaches the performance level of other methods' seventh active learning cycle by the end of the fifth cycle, and continues to improve thereafter. Similarly, for  CIFAR-100, EDALSSC attains the seventh-cycle performance of the baselines after six cycles. For SVHN and Fashion-MNIST, EDALSSC outperforms the best classification performance of most baselines after two and three cycles respectively.
Moreover, the learning curves of our method exhibit a relatively smooth and steady upward trend overall, that is to say,  EDALSSC achieves continuous and stable improvements between cycles. This suggests that our sample selection strategy effectively identifies the truly high-value samples for the current model. By prioritizing the selection of highly informative unlabeled samples that are most beneficial for model training, EDALSSC significantly enhances sample efficiency and accelerates performance gains. 


\begin{figure}[htbp]
  \centering
  \subfigure{
    \includegraphics[width=0.3\textwidth]{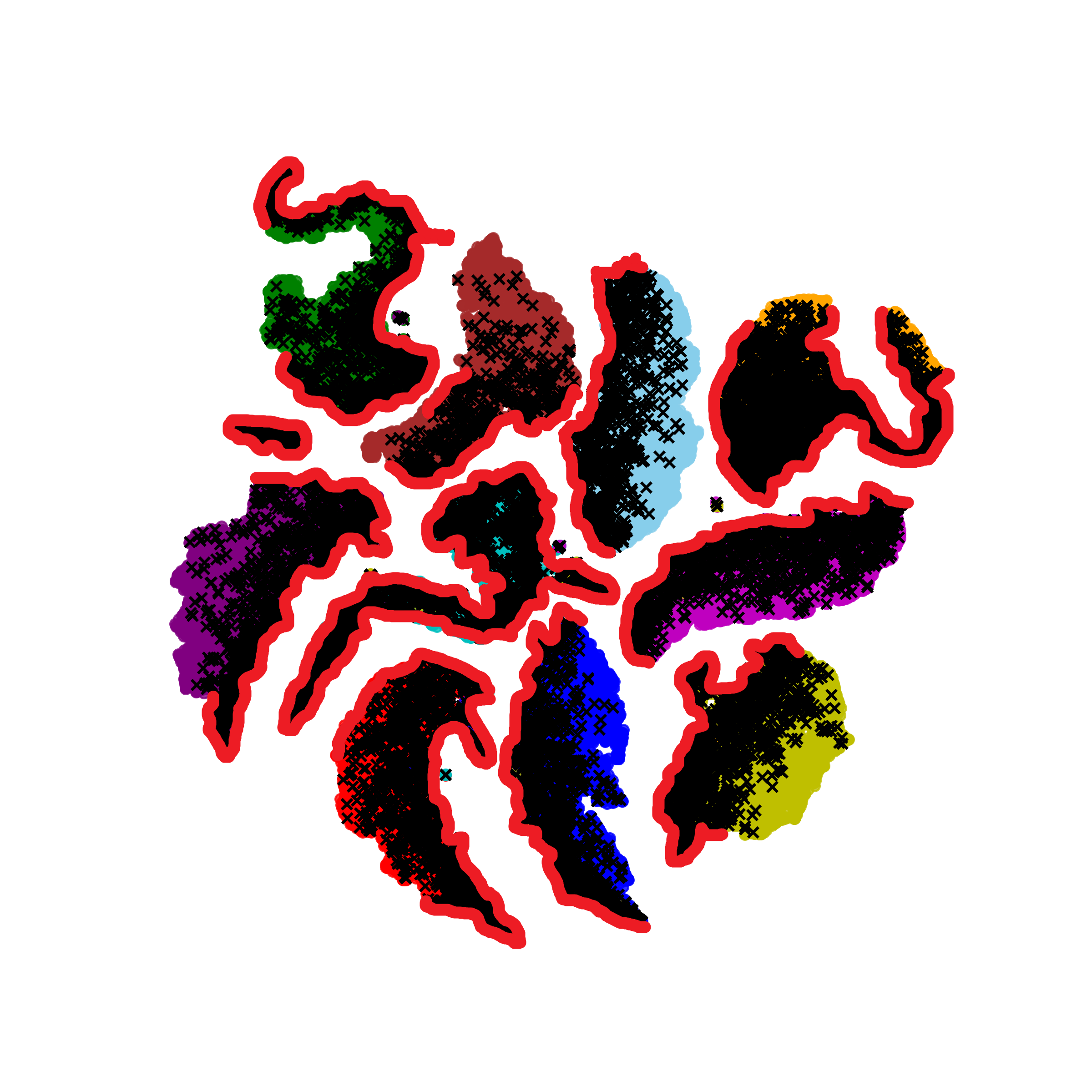}
    \label{fig:cifar10}
  }
  \subfigure{
    \includegraphics[width=0.3\textwidth]{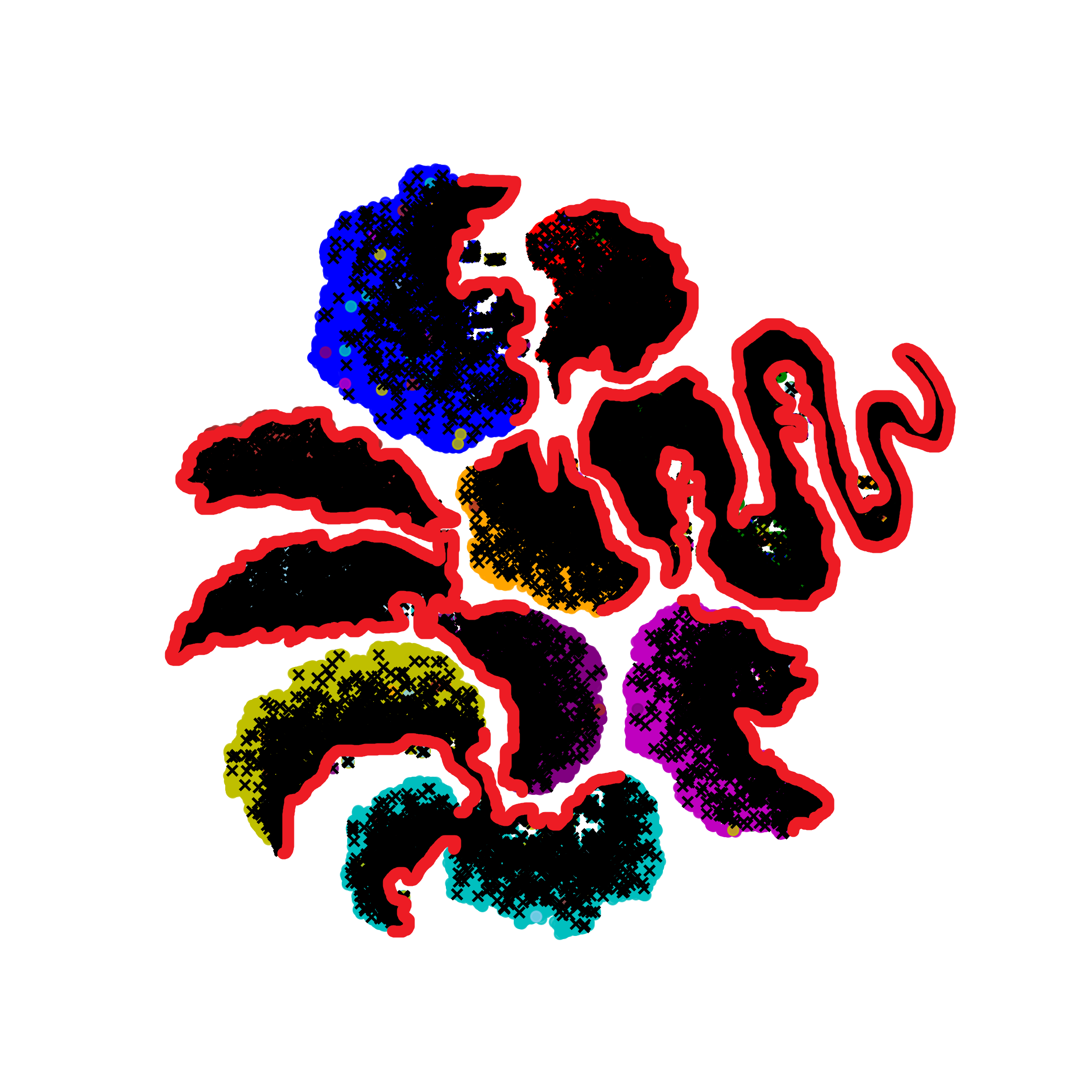}
    \label{fig:fashionMNIST}
  }
  \subfigure{
    \includegraphics[width=0.3\textwidth]{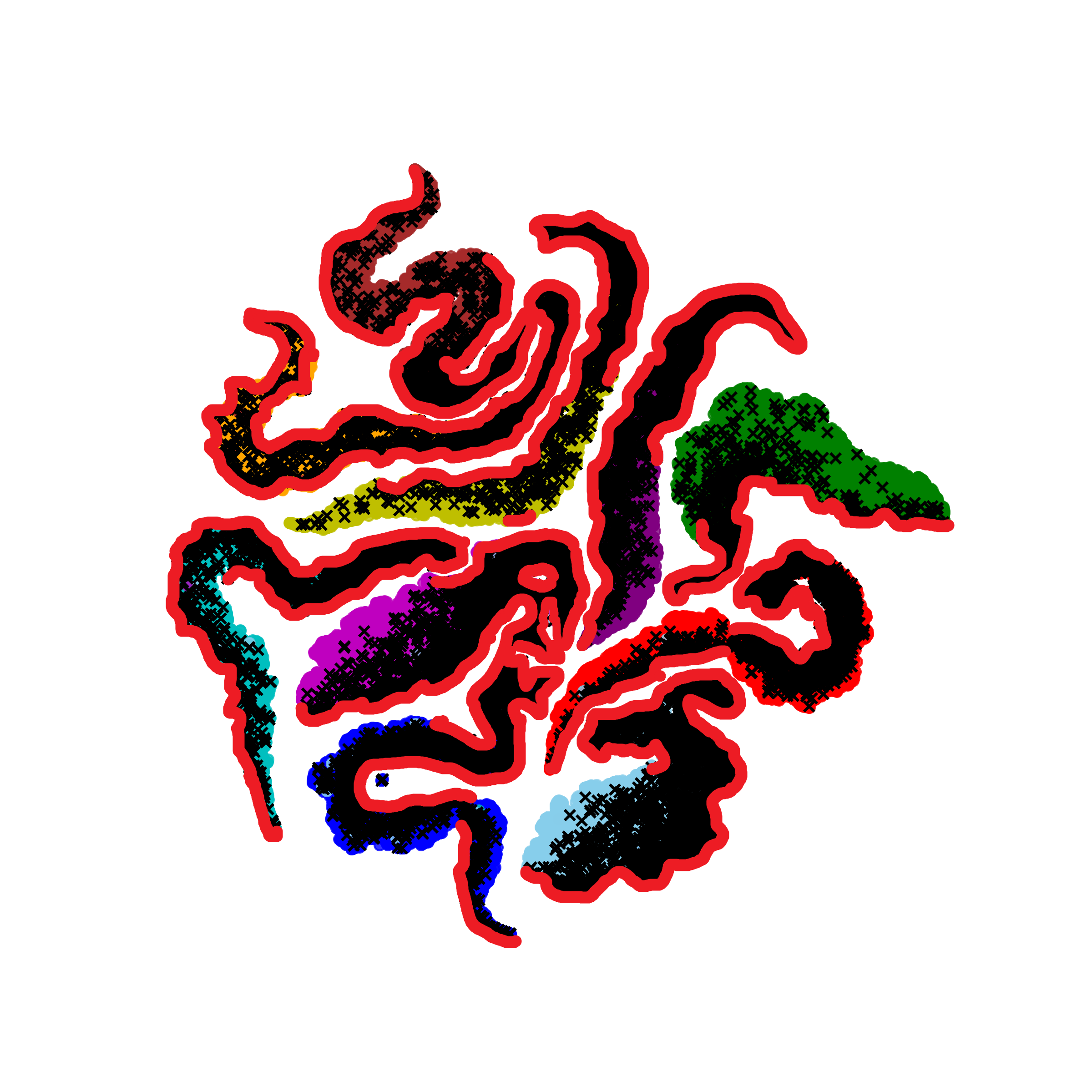}
    \label{fig:svhn}
  }
  
  \caption{The t-SNE visualization of the CIFAR10, SVHN and Fashion-MNIST dataset, highlighting the sample selection behavior of EDALSSC. Selected samples are marked in black, while unlabeled samples are shown in color. The boundaries of the class are marked with red lines.}
  \label{fig:per_class_accuracy}
\end{figure}

Furthermore, the full ablation study reveals that relying solely on the uncertainty estimation within the EDL framework leads to suboptimal performance. In contrast, EDALSSC framework substantially enhances active learning effectiveness by jointly modeling both ignorance information and conflict information to obtain a more robust uncertainty estimation. Additionally, Figure \ref{fig:per_class_accuracy} presents a t-SNE visualization of the selected samples within the feature space of the entire dataset. The selected points by EDALSSC are predominantly located near class boundaries, highlighting their informativeness and critical contribution to model improvement. This suggests that our selection strategy effectively identifies high-value samples that are essential for refining decision boundaries and enhancing overall performance.

\subsection{Ablation Study}


To rigorously assess the contribution of each component in  EDALSSC framework, this subsection conduct a series of ablation studies. Specifically, (1) Ablation 1 removes the uncertainty estimation module from the learning strategy. (2) Ablation 2 removes the dynamic scaling mechanism applied to the Dirichlet distribution parameter $\alpha$. (3) Ablation 3 disables the conflict information. (4) Ablation 4 defers uncertainty estimation until after model training. Each ablation is designed to isolate and evaluate the impact of a specific component introduced in EDALSSC.

The results of the ablation experiment are shown in Table \ref{tab:ablation-study}. It can be seen from the table that EDALSSC consistently outperforms all four ablation variants after seven iterations, achieving the highest classification accuracy rate. These results highlight the effectiveness of our core contributions, including: incorporating uncertainty estimation for unlabeled samples within the learning criterion; introducing a dynamic scaling mechanism for $\alpha$; Aggregating of conflict information and ignorance information based on model predictions; and computing uncertainty via loss fluctuations during training. Collectively, these components significantly enhance both sample selection efficiency and overall model performance.


\begin{table}[ht]
  \caption{Ablation Studies on CIFAR-10, CIFAR-100 and SVHN.}
  \label{tab:ablation-study}
  \centering
  \begin{tabular}{llccccccc}
    \toprule
    Dataset & Method & 1 & 2 & 3 & 4 & 5 & 6 & 7 \\
    \midrule
    \multirow{5}{*}{CIFAR-10} 
    & \textbf{EDALSSC} & 78.57 & 90.37 & 92.91 & 94.37 & 94.84 & 95.02 & \textbf{95.43} \\
    & Ablation 1 & 80.34 & 90.41 & 93.13 & 94.24 & 93.91 & 94.40 & 94.69 \\
    & Ablation 2 & 79.24 & 90.35 & 93.24 & 94.29 & 94.73 & 94.79 & 94.63 \\
    & Ablation 3 & 77.81 & 89.00 & 92.23 & 93.07 & 93.14 & 94.09 & 93.05 \\
    & Ablation 4 & 79.13 & 90.11 & 92.58 & 93.91 & 94.36 & 94.61 & 94.69 \\

    \midrule
    \multirow{5}{*}{CIFAR-100} 
    & \textbf{EDALSSC} & 29.12 & 51.00 & 59.88 & 64.11 & 66.86 & 68.72 & \textbf{70.60} \\
    & Ablation 1 & 0.64 & 50.57 & 59.34 & 63.11 & 65.71 & 67.91 & 69.36 \\
    & Ablation 2 & 28.87 & 51.14 & 59.25 & 63.16 & 66.24 & 68.83 & 70.19 \\
    & Ablation 3 & 29.76 & 50.21 & 58.49 & 62.41 & 64.07 & 65.31 & 67.35 \\
    & Ablation 4 & 29.19 & 50.00 & 57.61 & 64.61 & 66.12 & 69.44 & 69.89 \\

    \midrule
    \multirow{5}{*}{SVHN} 
    & \textbf{EDALSSC} & 93.64 & 96.45 & 96.76 & 96.96 & 96.76 & 96.93 & \textbf{97.08} \\
    & Ablation 1 & 93.81 & 96.53 & 96.85 & 96.78 & 96.99 & 96.68 & 78.87 \\
    & Ablation 2 & 93.82 & 96.58 & 96.77 & 96.79 & 96.79 & 49.76 & 64.67 \\
    & Ablation 3 & 94.27 & 96.25 & 96.48 & 96.81 & 95.80 & 95.91 & 96.46 \\
    & Ablation 4 & 93.85 & 96.58 & 96.82 & 96.93 & 96.89 & 96.92 & 96.90 \\

    \bottomrule
  \end{tabular}
\end{table}

\section{Conclusion}
\label{Conclusion}
This article proposes an evidential deep active learning framework for semi supervised classification tasks. EDALSSC introduces an uncertainty estimation mechanism to address the problem of traditional semi-supervised active learning methods lacking effective uncertainty estimation in model training and sample selection, which can lead to model overconfidence and difficulty in identifying high-value samples. In terms of learning strategy design, for labeled samples, the framework adopts evidence-based cross entropy loss,  where for unlabeled samples, the T-conorm operator is used to aggregate ignorance information and conflict information, thereby more comprehensively measuring the uncertainty estimation of the sample. In addition, EDALSSC has designed a dynamic adjustment mechanism for Dirichlet distribution parameters to balance the impact of number of classes and evidence strength on ignorance and conflict. In terms of sample selection strategy,  EDALSSC gives priority to the unlabeled samples with the greatest sum of uncertainty estimates in the later stage of training and when the training loss is increase, so as to  ensure the  high-value of the selected samples
The experimental results on image classification datasets such as CIFAR-10, CIFAR-100, SVHN, and Fashion-MNIST show that EDALSSC performs significantly better than existing advanced methods. Further ablation experiments also confirmed the crucial role of each component of EDALSSC in improving the overall model performance.


\small
\bibliographystyle{plain}
\bibliography{mybibfile}

\begin{thebibliography}{10}

\bibitem{barhoumi2025empirical}
Samia Barhoumi, Imene~Khanfir Kallel, {\'E}loi Boss{\'e}, and Basel Solaiman.
\newblock An empirical survey-type analysis of uncertainty measures for the
  fusion of crisp and fuzzy bodies of evidence.
\newblock {\em Information Fusion}, 121:103106, 2025.

\bibitem{bryan2005active}
Brent Bryan, Robert~C Nichol, Christopher~R Genovese, Jeff Schneider,
  Christopher~J Miller, and Larry Wasserman.
\newblock Active learning for identifying function threshold boundaries.
\newblock {\em Advances in neural information processing systems}, 18, 2005.

\bibitem{budd2021survey}
Samuel Budd, Emma~C Robinson, and Bernhard Kainz.
\newblock A survey on active learning and human-in-the-loop deep learning for
  medical image analysis.
\newblock {\em Medical image analysis}, 71:102062, 2021.

\bibitem{caramalau2021sequential}
Razvan Caramalau, Binod Bhattarai, and Tae-Kyun Kim.
\newblock Sequential graph convolutional network for active learning.
\newblock In {\em Proceedings of the IEEE/CVF conference on computer vision and
  pattern recognition}, pages 9583--9592, 2021.

\bibitem{chen2024r}
Mengyuan Chen, Junyu Gao, and Changsheng Xu.
\newblock R-edl: Relaxing nonessential settings of evidential deep learning.
\newblock In {\em The Twelfth International Conference on Learning
  Representations}, 2024.

\bibitem{gao2020consistency}
Mingfei Gao, Zizhao Zhang, Guo Yu, Sercan~{\"O} Ar{\i}k, Larry~S Davis, and
  Tomas Pfister.
\newblock Consistency-based semi-supervised active learning: Towards minimizing
  labeling cost.
\newblock In {\em European Conference on Computer Vision}, pages 510--526.
  Springer, 2020.

\bibitem{goodfellow2016deep}
Ian Goodfellow, Yoshua Bengio, Aaron Courville, and Yoshua Bengio.
\newblock {\em Deep learning}, volume~1.
\newblock MIT press Cambridge, 2016.

\bibitem{guo2021semi}
Jiannan Guo, Haochen Shi, Yangyang Kang, Kun Kuang, Siliang Tang, Zhuoren
  Jiang, Changlong Sun, Fei Wu, and Yueting Zhuang.
\newblock Semi-supervised active learning for semi-supervised models: Exploit
  adversarial examples with graph-based virtual labels.
\newblock In {\em Proceedings of the IEEE/CVF International Conference on
  Computer Vision}, pages 2896--2905, 2021.

\bibitem{he2016deep}
Kaiming He, Xiangyu Zhang, Shaoqing Ren, and Jian Sun.
\newblock Deep residual learning for image recognition.
\newblock In {\em Proceedings of the IEEE conference on computer vision and
  pattern recognition}, pages 770--778, 2016.

\bibitem{hekimoglu2024monocular}
Aral Hekimoglu, Michael Schmidt, and Alvaro Marcos-Ramiro.
\newblock Monocular 3d object detection with lidar guided semi supervised
  active learning.
\newblock In {\em Proceedings of the IEEE/CVF Winter Conference on Applications
  of Computer Vision}, pages 2346--2355, 2024.

\bibitem{huang2021semi}
Siyu Huang, Tianyang Wang, Haoyi Xiong, Jun Huan, and Dejing Dou.
\newblock Semi-supervised active learning with temporal output discrepancy.
\newblock In {\em Proceedings of the IEEE/CVF International Conference on
  Computer Vision}, pages 3447--3456, 2021.

\bibitem{krizhevsky2009learning}
Alex Krizhevsky, Geoffrey Hinton, et~al.
\newblock Learning multiple layers of features from tiny images.
\newblock 2009.

\bibitem{laine2016temporal}
Samuli Laine and Timo Aila.
\newblock Temporal ensembling for semi-supervised learning.
\newblock {\em arXiv preprint arXiv:1610.02242}, 2016.

\bibitem{lecun2015deep}
Yann LeCun, Yoshua Bengio, and Geoffrey Hinton.
\newblock Deep learning.
\newblock {\em nature}, 521(7553):436--444, 2015.

\bibitem{netzer2011reading}
Yuval Netzer, Tao Wang, Adam Coates, Alessandro Bissacco, Baolin Wu, Andrew~Y
  Ng, et~al.
\newblock Reading digits in natural images with unsupervised feature learning.
\newblock In {\em NIPS workshop on deep learning and unsupervised feature
  learning}, volume 2011, page~4. Granada, 2011.

\bibitem{ren2021survey}
Pengzhen Ren, Yun Xiao, Xiaojun Chang, Po-Yao Huang, Zhihui Li, Brij~B Gupta,
  Xiaojiang Chen, and Xin Wang.
\newblock A survey of deep active learning.
\newblock {\em ACM computing surveys (CSUR)}, 54(9):1--40, 2021.

\bibitem{sener2017active}
Ozan Sener and Silvio Savarese.
\newblock Active learning for convolutional neural networks: A core-set
  approach.
\newblock {\em arXiv preprint arXiv:1708.00489}, 2017.

\bibitem{sensoy2018evidential}
Murat Sensoy, Lance Kaplan, and Melih Kandemir.
\newblock Evidential deep learning to quantify classification uncertainty.
\newblock {\em Advances in neural information processing systems}, 31, 2018.

\bibitem{settles2009active}
Burr Settles.
\newblock Active learning literature survey.
\newblock 2009.

\bibitem{sinha2019variational}
Samarth Sinha, Sayna Ebrahimi, and Trevor Darrell.
\newblock Variational adversarial active learning.
\newblock In {\em Proceedings of the IEEE/CVF international conference on
  computer vision}, pages 5972--5981, 2019.

\bibitem{tarvainen2017mean}
Antti Tarvainen and Harri Valpola.
\newblock Mean teachers are better role models: Weight-averaged consistency
  targets improve semi-supervised deep learning results.
\newblock {\em Advances in neural information processing systems}, 30, 2017.

\bibitem{van2020survey}
Jesper~E Van~Engelen and Holger~H Hoos.
\newblock A survey on semi-supervised learning.
\newblock {\em Machine learning}, 109(2):373--440, 2020.

\bibitem{xiao2017fashion}
Han Xiao, Kashif Rasul, and Roland Vollgraf.
\newblock Fashion-mnist: a novel image dataset for benchmarking machine
  learning algorithms.
\newblock {\em arXiv preprint arXiv:1708.07747}, 2017.

\bibitem{yoo2019learning}
Donggeun Yoo and In~So Kweon.
\newblock Learning loss for active learning.
\newblock In {\em Proceedings of the IEEE/CVF conference on computer vision and
  pattern recognition}, pages 93--102, 2019.

\bibitem{zhang2020state}
Beichen Zhang, Liang Li, Shijie Yang, Shuhui Wang, Zheng-Jun Zha, and Qingming
  Huang.
\newblock State-relabeling adversarial active learning.
\newblock In {\em Proceedings of the IEEE/CVF conference on computer vision and
  pattern recognition}, pages 8756--8765, 2020.

\bibitem{zhang2022boostmis}
Wenqiao Zhang, Lei Zhu, James Hallinan, Shengyu Zhang, Andrew Makmur, Qingpeng
  Cai, and Beng~Chin Ooi.
\newblock Boostmis: Boosting medical image semi-supervised learning with
  adaptive pseudo labeling and informative active annotation.
\newblock In {\em Proceedings of the IEEE/CVF conference on computer vision and
  pattern recognition}, pages 20666--20676, 2022.

\bibitem{zhou2022modeling}
Qianli Zhou, {\'E}loi Boss{\'e}, and Yong Deng.
\newblock Modeling belief propensity degree: measures of evenness and diversity
  of belief functions.
\newblock {\em IEEE Transactions on Systems, Man, and Cybernetics: Systems},
  53(5):2851--2862, 2022.

\bibitem{zhou2023information}
Qianli Zhou, Witold Pedrycz, Yingying Liang, and Yong Deng.
\newblock Information granule based uncertainty measure of fuzzy evidential
  distribution.
\newblock {\em IEEE Transactions on Fuzzy Systems}, 31(12):4385--4396, 2023.

\bibitem{zhou2013active}
Shusen Zhou, Qingcai Chen, and Xiaolong Wang.
\newblock Active deep learning method for semi-supervised sentiment
  classification.
\newblock {\em Neurocomputing}, 120:536--546, 2013.

\bibitem{zhu2003combining}
Xiaojin Zhu, John Lafferty, and Zoubin Ghahramani.
\newblock Combining active learning and semi-supervised learning using gaussian
  fields and harmonic functions.
\newblock In {\em ICML 2003 workshop on the continuum from labeled to unlabeled
  data in machine learning and data mining}, volume~3, pages 58--65, 2003.

\bibitem{zhu2005semi}
Xiaojin~Jerry Zhu.
\newblock Semi-supervised learning literature survey.
\newblock 2005.

\end{thebibliography}
\normalsize

\end{document}